# A zone-based training approach for last-mile routing using Graph Neural Networks and Pointer Networks

## Authors
Àngel Ruiz-Fas[1], Carlos Granell[1], José Francisco Ramos[1], Joaquín Huerta[1], Sergio Trilles[1]

[1] *Institute of New Imaging Technologies, Universitat Jaume I, Avda. Sos Banyat, s/n, 12006, Castelló de la Plana, Spain*

## Abstract
Rapid e-commerce growth has pushed last-mile delivery networks to their limits, where small routing gains translate into lower costs, faster service, and fewer emissions. Classical heuristics struggle to adapt when travel times are highly asymmetric (e.g., one-way streets, congestion). A deep learning-based approach to the last-mile routing problem is presented to generate geographical zones composed of stop sequences to minimize last-mile delivery times.

The presented approach is an encoder–decoder architecture. Each route is represented as a complete directed graph whose nodes are stops and whose edge weights are asymmetric travel times. A Graph Neural Network encoder produces node embeddings that captures the spatial relationships between stops. A Pointer Network decoder then takes the embeddings and the route's start node to sequentially select the next stops, assigning a probability to each unvisited node as the next destination.

Cells of a Discrete Global Grid System which contain route stops in the training data are obtained and clustered to generate geographical zones of similar size in which the process of training and inference are divided. Subsequently, a different instance of the model is trained per zone only considering the stops of the training routes which are included in that zone.

This approach is evaluated using the Los Angeles routes from the 2021 Amazon Last Mile Routing Challenge. Results from general and zone-based training are compared, showing a reduction in the average predicted route length in the zone-based training compared to the general training. The performance improvement of the zone-based approach becomes more pronounced as the number of stops per route increases.

## 1. Introduction

E-commerce has maintained a rapidly growing trend in recent years [1]. Moreover, forecasts for the coming years show that this increase will continue. E-commerce faces many technical, organizational and logistics challenges, one of the most significant being the delivery of goods to customers at fixed locations (e.g., homes or lockers) in the most efficient way possible. This is known as the *last-mile routing problem*.

Last-mile routing is closely related to the Traveling Salesman Problem (TSP) [2]. This problem, extensively studied in the literature, has its mathematical roots in the 19th century and was initially examined by the mathematicians Sir William Rowan Hamilton and Thomas Penyngton Kirkman. The objective is to determine the most efficient route to visit a specific set of cities exactly once and return to the origin, thereby minimising the total travel cost. More precisely, last-mile routing is often modelled as a Vehicle Routing Problem (VRP) [3], which is a generalisation of the TSP. The VRP refers to the task of supplying geographically dispersed customers using a fleet of vehicles operating from a common depot. It has been proven to be a combinatorial optimization problem of NP-complete class.

In parallel, Artificial Intelligence (AI), and Machine Learning (ML) in particular, have experienced a rapid growth in recent years [4]. These advancements have allowed researchers to address problems that were previously too computationally complex to solve [5]. Central to these advancements are Neural Networks (NN), which constitute a core component of Deep Learning (DL) methods and drive many new approaches currently adopted by researchers worldwide [6].

In this work, the last-mile routing problem is addressed by combining NNs with a geospatial modelling perspective. A comparative analysis between a single model trained on the entire dataset and zone-specific models trained on partitioned geographical areas, considered as different instances of a model per generated zone, is performed to evaluate whether zone-specific training outperforms the single model. While the general model generates the complete route directly, the zone-trained models produce local (zone-level) sub-routes that are subsequently combined to form the complete route. The objective is to determine which strategy approximates ground-truth routes more accurately and optimizes delivery costs.

The rest of the article is organised as follows. Section 2 reviews related works on solving the last-mile routing problem. Section 3 describes the proposed approach and methodology. Section 4 introduces the experimental setup for the study. Section 5 presents and discusses the obtained results. Finally, Section 6 concludes the paper and outlines future work directions.

## 2. Related Works

In the literature, two main families of algorithms have been widely employed to address the last-mile routing problem: heuristic approaches and ML approaches. This section reviews studies using these different technological approaches.

### 2.1 Heuristic approaches

Most studies use heuristics to seek a solution that approximate the global optimal as closely as possible. Since the TSP is an NP-complete problem, finding the optimal solution is often impossible computationally. One of the most successful heuristic algorithms is the Lin-Kernighan-Helsgaun method (LKH) [7], which has been further improved in its third version (LKH3). This method is an improvement proposed by Helsgaun to the original Lin-Kernighan algorithm. The original Lin-Kernighan method [8] is based on the $\lambda$-opt algorithm, where, at each step of the algorithm, $\lambda$ edges of the tour are replaced to obtain a shorter route. Starting from a feasible tour, a $\lambda$-opt method improves the route in each step until it gets to a tour which is $\lambda$-optimal, this is, it is impossible to obtain a shorter tour by replacing any $\lambda$ of its links with any other set of $\lambda$ links.

Lin and Kernighan improved the $\lambda$-opt algorithm by introducing a variable $\lambda$. At each step, starting from $\lambda=2$, the algorithm checks whether a $\lambda+1$ exchanges can be considered. If so, it considers the new $\lambda$ value to perform the exchanges. Additionally, several heuristic rules are applied to limit the search space of the links to be replaced.

The improvement introduced by Helsgaun consists of using a measure called $\alpha$-nearness to select the edges to replace [7]. This measure is computed using minimum spanning 1-trees. If $T$ is a minimum 1-tree of length $L(T)$, and $T+(i,j)$ is a minimum 1-tree required to contain the edge $(i,j)$, the $\alpha$-nearness of edge $(i,j)$ is $\alpha(i,j)=L(T+(i,j))-L(T)$. That means that, given the length of any minimum 1-tree, the $\alpha$-nearness of an edge is the increase in length when the minimum 1-tree is required to contain that edge.

Wu and colleagues [9] applied a probabilistic model known as Prediction by Partial Matching (PPM) to train a high-order Markov model. By using the Rollout algorithm, they order the sequence of geographical zones containing delivery stops. Subsequently, the authors employed LKH3 to determine the optimal order of stops within each zone. This is a solution given to the Amazon Last Mile Routing Challenge [10], which is a challenge hosted by Amazon Last Mile Research team and supported by Massachusetts Institute of Technology's Center for Transportation Logistics. The objective of this competition was to get as close as possible to Amazon ground-truth routes. The winning team developed an improved version of LKH3, named LKH-AMZ [11], which added specific constraints and penalty functions to guide the edge replacement mechanism.

### 2.2 Machine Learning approaches

Compared to traditional heuristics, ML methods are less prevalent in the existing literature. Here, the focus is on two significant ML-based studies. The first study proposed an unsupervised encoder-decoder architecture to generate last-mile routes [12]. The encoder they use is similar to a Transformer architecture with attention layers composed of two sublayers: a Multi-Head Attention (MHA) layer and a fully-connected Feed Forward (FF) layer. With this encoder, they obtained an embedding for each node. Subsequently, an attention-based decoder compute, in each step, the probabilities of transitioning from the current node to each one of the other nodes. For that, log-probabilities (logits) are calculated for each node and passed through a softmax function to obtain normalised probabilities. The model is trained by optimising the loss function by gradient descent. The chosen gradient estimator was REINFORCE, using the expected tour length $L(\pi)$ and a baseline $b(s)$. The authors compared the results obtained by their model with those produced by the LKH3 solver. The performance gap between this model and LKH3 was close to 0, indicating the effectiveness of the proposed unsupervised learning approach.

In contrast to the previous work, the second study [13] used a supervised learning method. The authors employed a Graph Neural Network (GNN) model trained with the Adaptative Moment (Adam) optimizer and a binary cross entropy loss function. The model is designed to predict a Boolean value that indicates whether an edge is part of the optimal route [13]. The model achieves a training accuracy of 80.16% after 2000 training epochs with 221 instances. On a test dataset of 2,048 instances, the model also obtains an accuracy of 80%.

## 3. Proposed method

The first step to address the VRP is to model the data to be fed into a model. In this case, delivery routes can be represented using a graph structure. Each stop is modelled as a node of the graph, containing the features of the stop. In this work, each node contains as features the distances to the closest and the farthest stop, the distance to the route centroid, a 16-dimensional positional encoding and a 16-dimensional embedding which represents the zone in which it is located. The zone identifier is provided in the dataset. Then, each pair of nodes of the graph can be connected using two weighted directed edges, each of them containing the estimated travel time between them. Recall that, given a pair of edges $(a, b)$, the travel time from $a$ to $b$ can be different from the travel time from $b$ to $a$.

Since the results from [12] were very close to LKH3, which is considered the best performing algorithm for routing problems, an encoder-decoder architecture to the last-mile routing problem is also applied. Moreover, the study in [13] showed that GNNs are great mechanisms to compute embeddings of nodes in a graph. For this reason, a GNN is used as an encoder in this architecture. As for the decoder, [12] proposed a multi-head attention model. These types of models are well known to yield great results, but they require greater computational resources. Since the GNN is also heavier than other encoders, a Pointer Network (PN) [14] is adopted instead as the decoder, as it provides a lighter alternative while still achieving strong performance. Finally, there are many methods to compute the loss of the network to train it. Following the results from [12], the REINFORCE loss is selected, which is an algorithm that belongs to the Reinforcement Learning (RL) paradigm.

To generate the zones, a Discrete Global Grid System (DGGS) is first utilised, specifically Uber's Hexagonal Hierarchical Spatial Index (H3) [15]. H3 tiles the world with hexagonal cells supporting 16 resolutions Each cell has one seventh of the area of the coarser resolution. Approximately, the finer cells are contained within a parent cell. A key feature of H3 is its hierarchical indexing, where each cell has a unique identifier, and a cell's identifier can be truncated to obtain the identifier of its ancestor cell at a coarser resolution. Then, the selected delivery area is divided into subdivisions or zones, each comprising multiple hexagonal cells.

Previous studies [9] showed that the division of a selected area into smaller zones and ordering the stops within each zone can potentially outperform a global ordering approach with all of the stops. For that reason, in this work, a different instance of the model is trained per zone. Each of those models generates a sub-route containing the stops within its associated zone. Finally, all the sub-routes combined generate the final delivery route.

## 4. Experimental setup

The proposed method is tested using the routes of the Amazon Last Mile Routing Challenge [10]. These routes are divided into five metropolitan areas of the United States. The study focuses exclusively on routes within the Los Angeles metropolitan area, which contains the largest number of routes. Specifically, the dataset contains 2,888 training routes and 1,626 test routes in this metropolitan area. To divide this area into zones, H3 hexagons at resolution 7 are used and clustered using k-Means algorithm, obtaining 57 spatial clusters of similar sizes.

The encoder-decoder architecture with a GNN and a PN is developed using PyTorch library [16], employing PyTorch Geometric [17] for the GNN component. For the encoder, three GATv2 layers [18] are stacked, which are an improved version of Graph Attention Network (GAT) [19]. In these components, each node attends to its neighbours by using its own representation as a query, thereby generating an updated node state. Additionally, edge attributes are considered during the update process. After each GATv2 layer, a normalisation layer is applied. Between the normalisation step and the following GATv2 layer, an Exponential Linear Unit (ELU) activation function is added, which is common in graph-based models, as well as a dropout operation to mitigate overfitting and enhance the network's generalisation capabilities.

The decoder receives as input the node embeddings generated by the encoder. For this component, a Gated Recurrent Unit (GRU) to compute a hidden state is used and it is updated for each node of the graph using the embeddings. This allows the model to track the global context of the graph and improve route construction. Given the starting node, the PN decides, in each step, the next node to be visited in the route. The PN updates the GRU state using the embeddings and subsequently computes the attention values through a hyperbolic tangent function. In this implementation, these values are obtained by applying two Fully Connected (FC) layers, with a Rectified Linear Unit (ReLU) activation function between them, to the output of the GRU and to the node embeddings, respectively. The output of the FC layers is normalised prior to computing the hyperbolic tangent. Figure 1 graphically displays the described architecture.

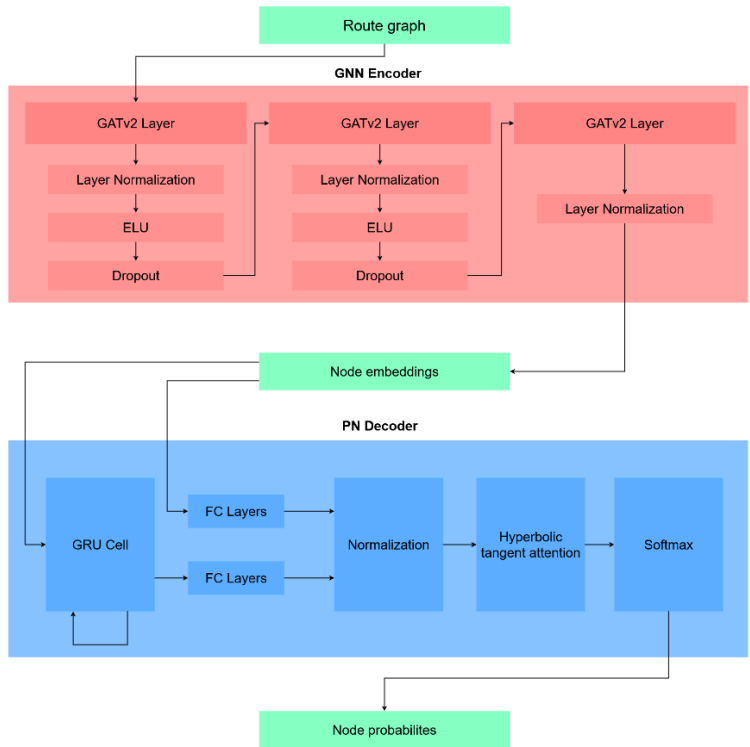

**Figure 1.** Proposed encoder–decoder architecture for last-mile routing, where a GNN-based encoder generates node embeddings that are sequentially processed by a PN decoder to produce routing decisions.

Experimental results are evaluated by computing the Mean Absolute Percentage Error (MAPE) between the ground-truth and predicted route lengths on the test fold. In addition, descriptive statistics of the MAPE are reported to provide deeper insight into the performance of both strategies. MAPE is chosen to facilitate the interpretation of error magnitude in relative terms.

The code of the model, alongside the training and inference scripts for reproducibility are available in the following GitHub repository: GitHub.

## 5. Results and discussion

When testing the trained model, the average route length obtained from the general training was 30,481.52, which represents a MAPE of 215.97% with respect to the ground-truth route lengths. Figure 2 displays the actual and predicted length for the first 150 routes in the test fold.

For zone-based training strategy, the average route length obtained on test samples was 21,973.6, corresponding to a MAPE of 134.57%. Figure 3 displays the actual and predicted route lengths for the first 150 routes of the test fold. Moreover, the prediction error for each route in the test fold is computed and the descriptive measures from these errors are extracted. Table 1 shows the error statistics obtained for both the general and the zone-based training strategies.

To study how the number of clusters visited in a route and the number of the stops in a route affect prediction quality, two different groupings of routes are performed. In the first case, routes are grouped by the number of clusters visited. In the second case, routes are grouped by number of stops. After that, the average actual route lengths for each group and the corresponding average predicted lengths are computed using both strategies. Moreover, the MAPE for each strategy within each group is calculated. Table 2 shows the results of the grouping by the number of clusters visited, while Table 3 shows the results of the grouping per number of stops.

Based on these results, it can be observed that models trained using the zone-based approach achieve lower errors than the general model. According to Table 1, both the average and the median errors are lower under the zone-based training. The minimum error is similar in both cases, although slightly lower for the zone-based strategy. The maximum error is significantly lower in zone-based training, with a difference of more than 50,000 s. Although these results remain far from the ground-truth routes, the improvement achieved by the zone-based approach is remarkable.

Regarding the quantiles shown in Table 1, all values are also lower under zone-based training, and the difference becomes more pronounced as the quantile increases. This observation, together with the reduction in maximum error, suggests that the main improvements achieved by zone-based training, compared to general training, occur in scenarios with higher errors. Table 3 shows that high errors typically correspond to longer routes with more stops. An important point to highlight is that the greater the number of stops in a route, the larger the geographical area it covers. Moreover, as the number of stops increases, the number of possible routes grows, resulting in more edges in the graph that can potentially be part of the final tour. The zone-based training strategy helps to partition these large areas, thereby reducing the highest errors by removing connections between stops that are geographically distant. In addition, the number of edges is reduced by splitting the main graph into more manageable subgraphs.

When the errors in the time predictions depending on the number of clusters visited and the number of stops per route are compared, it is possible to observe some interesting behaviours. As shown in Table 2, when varying the number of traversed clusters, the MAPE of the general model remains relatively stable, as expected, given that the model does not change and performs inference independently of the number of clusters. However, a clear decrease in the MAPE of the zone-based training strategy is observed as the number of clusters increases. This behaviour corresponds with the findings discussed in the previous paragraph, where zone-based training was shown to limit the search space of possible routes when they cover a large area, reducing the number of potential paths and, consequently, improving prediction accuracy. Moreover, Table 2 confirms that the routes tend to be longer in time as the number of clusters increases.

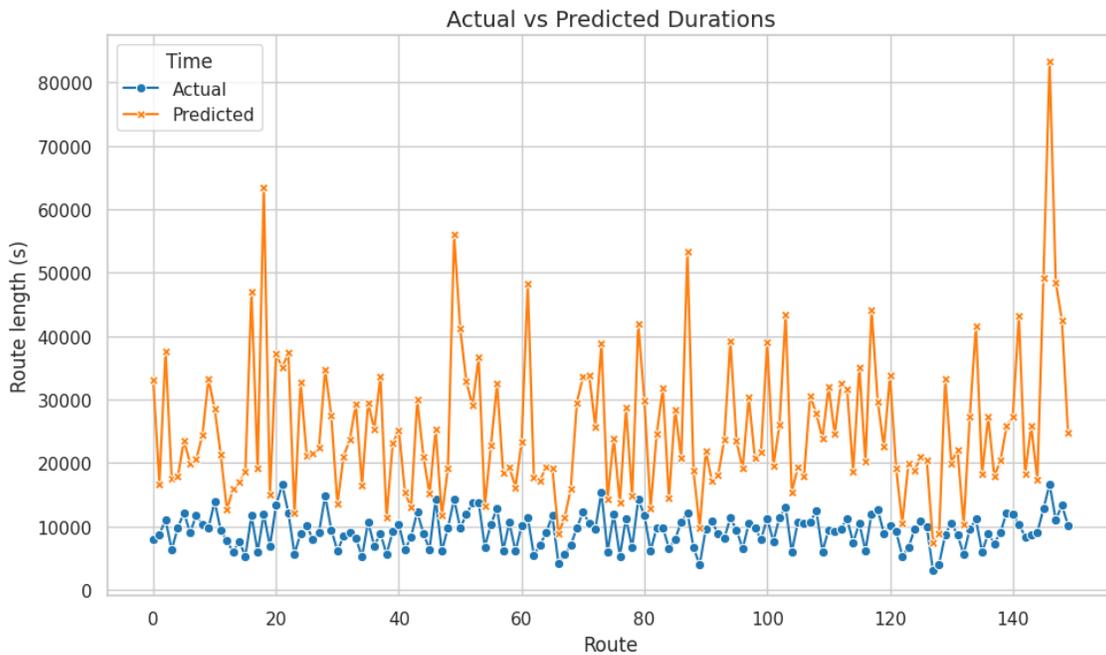

**Figure 2. Actual and predicted route lengths in general training strategy.**

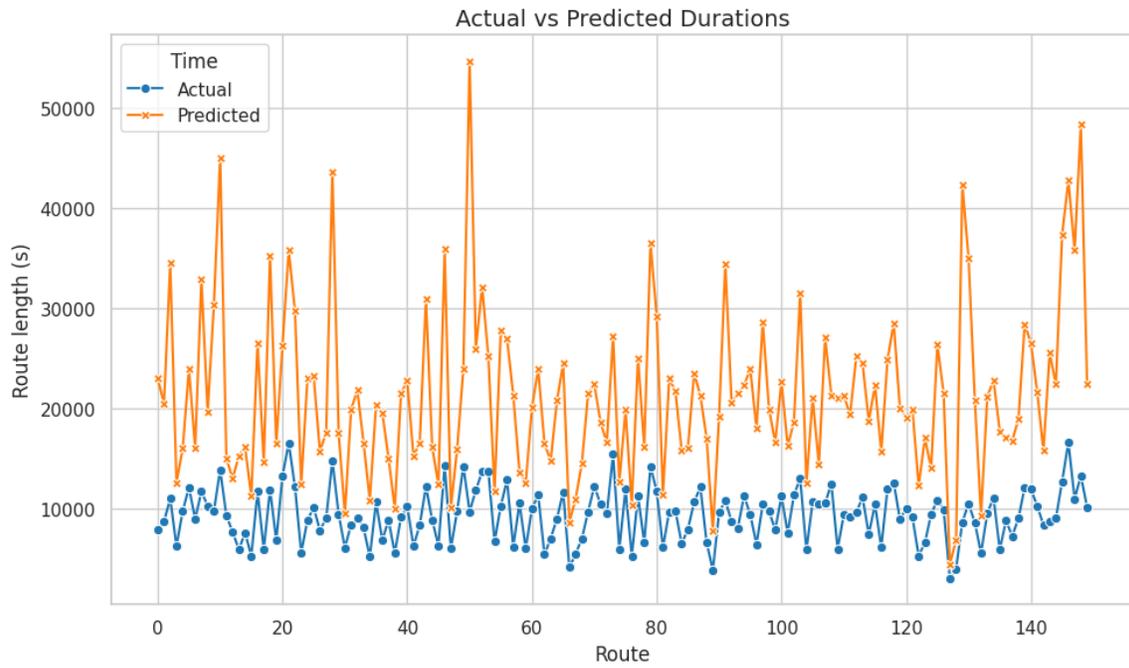

Figure 3. Actual and predicted route lengths in zone-based training strategy.

| Error metric / Strategy | General | Zone-based |
|---|---|---|
| Average error | 21,128.85 | 12,620.93 |
| Maximum error | 108,299.9 | 53,103.3 |
| Minimum error | 1,369.5 | 1,278.4 |
| Error quantile 0.25 | 11,821.87 | 8,118.85 |
| Error median (quantile 0.5) | 18,340.25 | 11,580.35 |
| Error quantile 0.75 | 26,784.55 | 15,494.87 |
| Error quantile 0.9 | 37,879.7 | 20,692.65 |

Table 1. Error statistics of both strategies in test fold.

| Metric / Clusters | One | Two | Three | Four+ |
|---|---|---|---|---|
| Mean actual length (s) | 9,366.6 | 8,843.22 | 10,386.35 | 11,817.75 |
| Mean predicted length (general) (s) | 32,939.99 | 27,768.22 | 35,747.64 | 42,414.41 |
| Mean predicted length (zone-based) (s) | 33,320.6 | 20,858.34 | 22,742.68 | 25,620.44 |
| General MAPE (%) | 250.67 | 204.8 | 234.96 | 255.07 |
| Zone-based MAPE (%) | 255.51 | 134.81 | 117.92 | 116.75 |

Table 2. Errors per number of clusters.

| Metric / Stops | ≤ 100 | 101 - 120 | 121 – 140 | 141 – 160 | 161 - 180 | 181 - 200 | ≥ 200 |
|---|---|---|---|---|---|---|---|
| Mean actual length (s) | 5,678.27 | 7,717.52 | 9,051.59 | 10,366.37 | 11,173.10 | 11,687.11 | 14,258.77 |
| Predicted (general) (s) | 13,307.44 | 21,032.57 | 28,936.65 | 34,544.20 | 41,005.00 | 42,368.89 | 59,771.28 |
| Predicted (zone-based) (s) | 11,186.86 | 16,518.21 | 20,606.55 | 24,868.91 | 27,868.01 | 30,663.61 | 39,316.88 |
| General MAPE (%) | 132.61 | 172.47 | 215.25 | 232.73 | 266.01 | 263.16 | 318.58 |
| Zone-based MAPE (%) | 98.16 | 118.49 | 132.35 | 142.44 | 152.90 | 163.64 | 183.32 |

**Table 3. Prediction errors grouped by number of route stops.**

In Table 3, it is possible to see that the length of the routes increases as the number of stops increases. The MAPE, both in general training and in zone-based training, also increases as the number of stops (and thus the route duration) grows. This behaviour can be caused by a larger search space. However, the zone-based training strategy consistently yields better results, with a lower MAPE in all cases and a smaller increase across the different numbers of stops of a route. Clearly, as the size of the route increases, the difference between the MAPEs of both approaches becomes larger, indicating that the model struggles more to find optimal routes as route complexity increases.

These results are subject to several limitations. The main limitation encountered was the hardware, particularly GPU VRAM, which constrained the choice of model hyperparameters. In addition, the dataset used presents certain constraints, as it does not contain extensive information for route prediction beyond stop coordinates, zone identifiers, and travel time estimations between stops. Additional information, such as traffic conditions or travel distance in meters between stops, would likely improve model performance.

## 6. Conclusion

This work presents an encoder-decoder architecture to solve the last-mile routing problem. The encoder is based on a GNN composed of three GATv2 layers, with normalisation layers between them, while the decoder relies on a PN with a GRU and the classical PN attention mechanism based on hyperbolic tangent function. Although the proposed model does not yet achieve accurate route predictions in absolute terms, with a MAPE above 100%, the results provide valuable insights into the behaviour of deep learning-based routing models under realistic constraints.

A clear tendency in the results can be observed, where prediction errors increase as the route lengths grow. This behaviour is consistent with the fact that most routes in the dataset contain around 150 stops, which represents a challenging scenario for last-mile routing. Consequently, the model performs more reliably on routes with smaller number of stops, as shown in Table 3, where the difference in MAPE between the shortest and longest routes is approximately 186% for general training and 85% for zone-based training.

The main objective of this work was to compare general training with zone-based training and to assess whether geographical partitioning can improve last-mile routing performance. Based on the experimental results, it can be concluded that the zone-based training consistently outperforms general training, particularly for longer routes. This improvement is reflected in a substantial reduction of the largest prediction errors, leading to lower overall MAPE values. For the longest routes, Table 1 shows a difference of approximately 135% between the MAPEs of general and zone-based training strategies, while Table 2 indicates a difference of around 139% for routes that span a larger number of clusters.

Despite these improvements, the results remain far from the ground-truth routes, and several limitations must be acknowledged. One of the most critical factors is the batch size used during training. A previous work [12], employs batch sizes of up to 512, which enables better gradient estimation and generalisation. In this case, hardware constraints, specifically GPU VRAM, limited the batch size to 32, which may have restricted the learning capacity of the model. Although a batch size of 64 was feasible, it required reducing the node embedding dimensionality and did not yield performance gains. Access to more powerful computational resources would likely enable larger batch sizes, longer training schedules, and improved model configurations, potentially leading to better solutions.

For future work, the main objective is to gain access to more powerful hardware capable of supporting larger models, thereby allowing the model to learn and generalise more effectively, in line with the articles reviewed in this work. In particular, the most relevant improvement enabled by better hardware would be an increase in the batch size, which is identified as a critical factor for reducing the MAPE of the proposed model. In addition, improved computational resources allow us to replace the PN decoder with a Multi-Head Attention (MHA) mechanism, which is expected to provide better performance than the current decoder architecture.

## Acknowledgements

This publication is part of the project PID2022-141813OB-I00 funded by MCIN/AEI/10.13039/501100011033 and by ERDF/EU. Project supported by a 2024 Leonardo Grant for Scientific Research and Cultural Creation from the BBVA Foundation.

## References

[1] 'Global retail e-commerce sales 2022-2028', Statista. Accessed: Dec. 09, 2025. [Online]. Available: https://www.statista.com/statistics/379046/worldwide-retail-e-commerce-sales/

[2] R. Matai, S. Singh, and M. Lal, 'Traveling Salesman Problem: an Overview of Applications, Formulations, and Solution Approaches', in *Traveling Salesman Problem, Theory and Applications*, D. Davendra, Ed., InTech, 2010. doi: 10.5772/12909.

[3] N. Christofides, 'The vehicle routing problem'.

[4] 'The state of AI in 2023: Generative AI's breakout year | McKinsey'. Accessed: Dec. 09, 2025. [Online]. Available: https://www.mckinsey.com/capabilities/quantumblack/our-insights/the-state-of-ai-in-2023-generative-ais-breakout-year

[5] S. Joksimovic, D. Ifenthaler, R. Marrone, M. De Laat, and G. Siemens, 'Opportunities of artificial intelligence for supporting complex problem-solving: Findings from a scoping review', *Comput. Educ. Artif. Intell.*, vol. 4, p. 100138, 2023, doi: 10.1016/j.caeai.2023.100138.

[6] 'What Is a Neural Network? | IBM'. Accessed: Dec. 09, 2025. [Online]. Available: https://www.ibm.com/think/topics/neural-networks

[7] K. Helsgaun, 'An effective implementation of the Lin–Kernighan traveling salesman heuristic', *Eur. J. Oper. Res.*, vol. 126, no. 1, pp. 106–130, Oct. 2000, doi: 10.1016/S0377-2217(99)00284-2.

[8] S. Lin and B. W. Kernighan, 'An Effective Heuristic Algorithm for the Traveling-Salesman Problem', *Oper. Res.*, vol. 21, no. 2, pp. 498–516, 1973.

[9] C. Wu, Y. Song, V. March, and E. Duthie, 'Learning from Drivers to Tackle the Amazon Last Mile Routing Research Challenge', 2022, *arXiv*. doi: 10.48550/ARXIV.2205.04001.

[10] D. Merchán *et al.*, '2021 Amazon Last Mile Routing Research Challenge: Data Set', *Transp. Sci.*, vol. 58, no. 1, pp. 8–11, Jan. 2024, doi: 10.1287/trsc.2022.1173.

[11] W. Cook, S. Held, and K. Helsgaun, 'Constrained Local Search for Last-Mile Routing', 2021, *arXiv*. doi: 10.48550/ARXIV.2112.15192.

[12] W. Kool, H. van Hoof, and M. Welling, 'Attention, Learn to Solve Routing Problems!', 2018, *arXiv*. doi: 10.48550/ARXIV.1803.08475.

[13] M. Prates, P. H. C. Avelar, H. Lemos, L. C. Lamb, and M. Y. Vardi, 'Learning to Solve NP-Complete Problems: A Graph Neural Network for Decision TSP', *Proc. AAAI Conf. Artif. Intell.*, vol. 33, no. 01, pp. 4731–4738, July 2019, doi: 10.1609/aaai.v33i01.33014731.

[14] O. Vinyals, M. Fortunato, and N. Jaitly, 'Pointer Networks', 2015, *arXiv*. doi: 10.48550/ARXIV.1506.03134.

[15] 'H3: Uber's Hexagonal Hierarchical Spatial Index', Uber Blog. Accessed: Dec. 10, 2025. [Online]. Available: https://www.uber.com/en-EG/blog/h3/

[16] A. Paszke *et al.*, 'PyTorch: An Imperative Style, High-Performance Deep Learning Library', 2019, *arXiv*. doi: 10.48550/ARXIV.1912.01703.

[17] M. Fey and J. E. Lenssen, 'Fast Graph Representation Learning with PyTorch Geometric', 2019, *arXiv*. doi: 10.48550/ARXIV.1903.02428.

[18] S. Brody, U. Alon, and E. Yahav, 'How Attentive are Graph Attention Networks?', 2021, *arXiv*. doi: 10.48550/ARXIV.2105.14491.

[19] P. Veličković, G. Cucurull, A. Casanova, A. Romero, P. Liò, and Y. Bengio, 'Graph Attention Networks', 2017, *arXiv*. doi: 10.48550/ARXIV.1710.10903.